# Prediction of Platinum Prices Using Dynamically Weighted Mixture of Experts

Baruch Lubinsky, Bekir Genc and Tshilidzi Marwala
University of the Witwatersrand
Private Bag x3
Wits, 2050, South Africa

*Abstract*—**Neural networks are powerful tools for classification and regression in static environments. This paper describes a technique for creating an ensemble of neural networks that adapts dynamically to changing conditions. The model separates the input space into four regions and each network is given a weight in each region based on its performance on samples from that region. The ensemble adapts dynamically by constantly adjusting these weights based on the networks' current performance. The dataset used is a collection of financial indicators with the goal of predicting the future platinum price. An ensemble with no weightings does not improve on the naive estimate of no weekly change; our weighting algorithm gives an average percentage error of 63 % for twenty weeks of prediction.**

## I. INTRODUCTION

Neural networks are known to be powerful tools for predicting future values of a time series [1]. Neural network regression is especially powerful in cases where simple linear regression is ineffective due to the non linear nature of the system [2]. One such system is the financial market, in which accurate predictions are difficult to make. This paper examines a neural network technique used to forecast the price of platinum.

This paper demonstrates the advantage gained by spatially dividing an input space. That is separating the input into different regions and giving each member of an ensemble a weight in each network. These weights are then extended to be functions of time in the study of platinum price predictions.

The approach taken gives positive results with a relatively high accuracy of prediction to this notoriously difficult problem [3]. A mixture of experts is used in order to cover the diverse factors that affect the future price of platinum. The voting weight allocated to each hypothesis is updated after each test sample. This dynamic weighting is a novel approach to the problem and is shown to greatly increase the accuracy of the ensemble.

## II. MIXTURE OF EXPERTS

The power of a neural network to make predictions can be greatly increased by combining the output of a number of networks collected in an ensemble [4]. For cases where the system is too complex to be learned by a single network, a mixture of experts can be used. Each network can correctly learn some feature of the system. These networks can then be combined to provide a model for the whole system. The method of combining the networks will depend on the nature of the data being modeled [5].

## III. SPATIAL DIVISION OF INPUT SPACE

One method of combining the outputs of the networks in an ensemble is to simply take the mean of all the outputs. However this does not take advantage of the fact that each network may have learned a different feature of the dataset. Any input space can be divided spatially along different features to create regions in the input space [6]. The performance of each expert can then be judged per region. Each network is assigned a numerical weight in each region. Then the output of the ensemble is the weighted average of the each network's output [7]. Consider an ensemble made up of networks $f_k$ with corresponding weights $w_k(region)$. For a given input $\vec{x}$ in region $i$ the prediction of the ensemble is:

$$y = \frac{\sum_k f_k(\vec{x}) w_k(i)}{\sum_k w_k(i)} \qquad (1)$$

Thus the contribution of the networks with the highest weights will have the greatest impact on the value of the output. The weights are initially defaulted to a value of 1. When all the weights are 1, the output is simply the mean of the output of each network.

### A. Preliminary Testing

This method of dividing the input space into different regions is tested on a sample dataset, created using PRTools [8]. Figure 1 shows an example of the "Banana" dataset generated by PRTools. The different markers represent the two classes to be classified.

The classifier used is an ensemble of multi-layer perceptron (MLP) neural networks. The dataset is divided into 150 and samples for training and 50 for testing. Using different data points for testing than training ensures that the generalization ability of the ensemble is tested and that the classifier does not over fit to the training samples. Figure 1 shows the different regions of the input space, separated by the dashed lines. The regions are created by separating both axes by the median of the feature. Each network has a vector of weights corresponding to each region. These weights are adjusted during training. For each sample, if the network classifies correctly, the relevant weight is multiplied by 1.2 otherwise it is multiplied by 0.4. These values are found to give weights that are constrained to reasonable values. When the ensemble is tested, the output is then the weighted average of the output of each network, according the weights calculated.

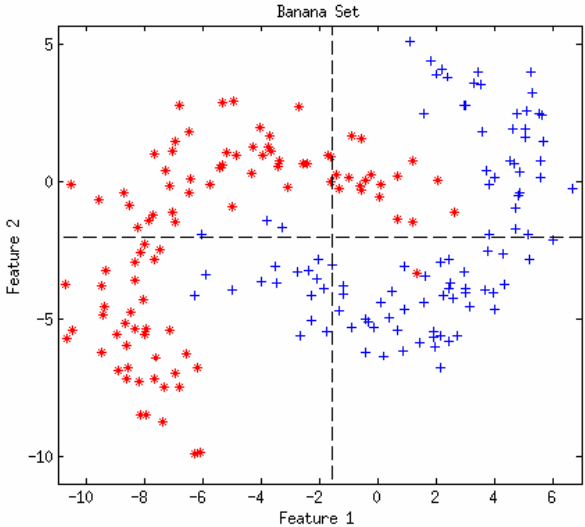

**Fig. 1. Data used to demonstrate the power of dividing the input space**

The accuracy of the ensemble is tested 100 times to obtain a meaningful average for four different weighting schemes. The test is run for networks with no weights (all weights are 1) and the decision is simply the mean of all the outputs. A small improvement is gained by giving each network one weight corresponding to its overall performance. Then the input space is divided into two regions by only dividing on one feature with two weights, and then four regions. The results are shown in table I, accuracy is the number of correct classifications over the total number of test samples.

TABLE I
CLASSIFIER ACCURACY WITH DIFFERENT WEIGHTINGS

| Test | Accuracy |
|---|---|
| No weights | 82.92 % |
| 1 weight | 83.50 % |
| 2 weights | 86.24 % |
| 4 weights | 88.18 % |

These results show that the performance of an ensemble is improved by giving more strength to the output of a network that has better accuracy. The performance of the ensemble is improved even further when the input space is divided and weights are assigned for each region. These regions need not divide the different classes perfectly to be effective. The regions in figure 1 are separated along the median of each feature which proves to be an adequate method for defining the regions. This test shows that the divisions in the input space need not represent any complex feature of the dataset. This relatively simple data demonstrates the power of weighting the different experts in an ensemble as a function of the position of the input.

IV. DYNAMIC WEIGHTING

The previous section describes an effective method of combining a number of neural networks into an ensemble with significantly improved performance. However, the data in that example – or more importantly the function that generates them – is stationary. We are interested in data being drawn from a dynamic environment. In this case, it is not sufficient to train the networks and combine them in an ensemble. The factors which generated the training data are unlikely to be present at the time of testing. The ensemble needs to adapt to changing conditions.

The concept of spatial weightings is extended for time series such that the weightings change as a function of time [9]. The weight for each region is updated after each sample. This is possible for time series in which, as each sample is received, the correct output for the previous sample becomes known. Adjusting the weights in this manner is a powerful method of implementing an adaptive model of a system. At each time step, the ensemble is updated without retraining each network. If we assume that the factors governing a system vary slowly within some bounded space, an ensemble with dynamic weights can retain its accuracy over time even as the system changes. Such a model can adapt continuously provided the conditions of the system were encountered in training. Thus effective adaptation is achieved, without the cost of retraining the ensemble.

*A. Platinum Price Case Study*

An example of a system that displays the characteristics described above is the platinum price. This is a notoriously difficult system to model due to the vast range of factors that impact it [3]. However, these factors are limited within a reasonable scope. The price of platinum is used here as an example to illustrate the power of dynamic weighting to make predictions in a complex system.

*1) Neural Network Structure:* Each expert in the ensemble is a MLP neural network. The network takes as its input the current market trends and the output is the prediction for the future change in platinum price. The inputs are the prices of platinum, palladium, radium, gold and Brent Crude and the South African Rand to US Dollar exchange rate as these are considered to be the best indicators [3]. The data is smoothed by taking weekly averages and then normalized by considering the percentage change for each week. The inputs are the changes during the previous week. The correct output is the percentage change in platinum price during the subsequent week.

Individual networks are trained by the Markov Chain Monte Carlo (MCMC) method [10]. The weights of the network are initialized randomly and then adjusted in small random steps in an attempt to reduce the mean square error over the set of training data. This ensures that the full weight space of the networks is explored and thus increases the diversity and generalization of the ensemble. The networks are found, heuristically, to perform best with two hidden nodes. The activation function at the hidden layer is a sigmoid, which is shifted to constrain the output on the range of [−1; 1]. On the output nodes, the activation function, also a sigmoid, ensures that the output is in the range [−0.2; 0.2] as this is the range of actual weekly price changes.

*2) Ensemble:* An ensemble is initially created with on network trained for a short amount of time – 10 epochs – on the full training dataset. This network becomes the initial benchmark for the ensemble. The accuracy of the ensemble is measured by taking the mean normalized square error over all the samples. If $t_n$ is the correct output for input $n$ and $y(n)$ is the output of the ensemble, the error is:

$$error = \frac{1}{N} \sum_n \left( \frac{y(n) - t_n}{t_n} \right)^2 \quad (2)$$

for $N$ data points. If $t_n = 0$ the square error is used. It is necessary to take the percentage error measurement so that an ensemble does not appear to be performing well simply because the price changes slowly and the absolute error is small.

An ensemble is trained on a full set of training data, 100 samples are sufficient to expose the ensemble to a large range of market forces. At each iteration, a new random network is created and trained on a subset of the training data – 20 weeks in this study. This gives the network a chance to become specialized on a small portion of the data. Then the network is added to the ensemble. If this decreases the error over the full training set, the network is retained, otherwise it is discarded. This method of selection leads to good generalization performance of the ensemble. Training continues in this manner for a fixed number of iterations or until some stopping criterion is met.

The input space is divided into four regions according to the two features which are considered most significant. The exchange rate and gold prices are used with the division along the zero line (which is close to the median for both features). Thus the sections are based on whether the value is increasing or decreasing. The specific selection of regions is not important, provided that the input space is divided fairly evenly. Each network within the ensemble is assigned a weight in each region corresponding to its performance on samples from that region. The output of the ensemble is the weighted average of the output of each of the experts.

The power of the ensemble comes from the way that the weights of each network are set and updated. Following training, an ensemble is considered to contain experts on the varied range of factors that impact the market. Initially each expert is weighted equally by a vector of 1's corresponding to each portion of the input space. In order to make accurate predictions, the weights of the networks must be adjusted for the current market situation. This method relies on the assumption that the factors influencing the price of platinum exist in a bounded space and vary slowly.

After each sample becomes known the weights are recalculated for the 10 previous samples. It is not necessary to retain the past input data, as each network's output will not change. The most recent sample is given the most significance as shown in figure 2.

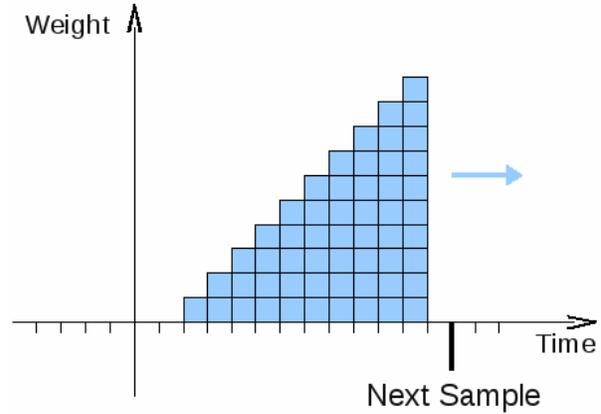

**Fig. 2. Decreasing contribution of older samples to the weight**

The weight in each region tends exponentially towards 1 so that if a weight – positive or negative – is not being continuously reinforced, it will lose its significance. This prevents any weight from dominating the ensemble after it has become irrelevant.

*3) Results:* The performance of an ensemble in predicting the future price of platinum is measured by equation 2. A naïve prediction would guess that there is no change in the price each week, leading to an error of 1. This is the benchmark against which any prediction is compared. A number of tests are run to compare the performance of different models. In each case, 10 ensembles are trained and tested; the average performance is given in table II.

TABLE II
PREDICTION ACCURACY WITH DIFFERENT WEIGHTING SCHEMES

|  | 4 weeks | 10 weeks | 20 weeks |
| --- | --- | --- | --- |
| Unweighted | 1.1374 | 1.1155 | 1.0265 |
| Static Weight | 0.8602 | 0.9708 | 1.024 |
| Dynamic Weight | 0.4461 | 0.5223 | 0.7416 |

The ensembles with constantly updated weights (Dynamic Weight) clearly outperform the ensembles which are un-weighted or statically weighted. The statically weighted ensembles are weighted at the start of the test period, but those weightings remain fixed. This gives an advantage in the short term, but over a longer time period does not improve the performance at all.

The results of table II are achieved by ensembles in which each network is trained for 20 epochs. Increasing this period to 40 epochs improves the performance of the dynamically weighted ensembles to 0.4069 over 11 weeks.

This corresponds to an average error of about 63 %, by taking the square root, which is considered good for data of this type [11]. The predicted change is plotted along with the actual change in figure 3.

This shows the accuracy of the predictions, which are close to the actual values and follow the shape well. The worst predictions are on samples with the largest change and in these cases, the predictor makes a smaller prediction in the correct direction. In addition to accuracy of prediction, it is significant that direction of the change is predicted correctly for each week in this example. This is a useful property for a financial predictor to have. The problem can be reduced to a binary classification of whether the price will increase or decrease during the following week. This model generally has an accuracy of about 90 % for the binary classification.

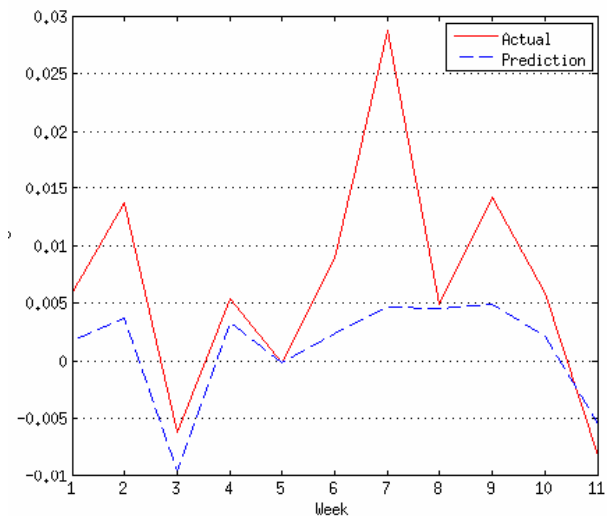

**Fig. 3. Predicted and actual weekly price change**

## V. CONCLUSION

Making predictions of commodity prices, such as platinum, is a difficult and attractive task. This neural network approach provides good predictions of the platinum price. This is achieved via a mixture of experts with dynamic weights. The weights attributed to each network are functions of both the region of the input and time. This is a novel approach which is well suited to this complex system, which changes over time. The ensemble is able to adapt without retraining which saves both computational time and memory. The results show that this is a powerful method of using neural networks for time series prediction.